\documentclass{article}

\usepackage{multirow}
\usepackage[para]{footmisc}
\usepackage[hyphens]{url}
\usepackage{lineno,hyperref}
\usepackage{breakurl}
\usepackage{amsmath}
\usepackage{color}
\usepackage{authblk}
\usepackage{graphicx}
\usepackage{natbib}
\usepackage{endnotes}

\graphicspath{ {img/} }

\let\footnote=\endnote

\begin{document}


\title{Authorship Attribution Using the Chaos Game Representation}

\author[1]{Daniel Lichtblau}
\author[2]{Catalin Stoean\thanks{Corresponding author.}}

\affil[1]{Wolfram Research, Champaign, Illinois, USA\\ E-mail: danl@wolfram.com}
\affil[2]{Faculty of Sciences, University of Craiova, Romania\\ Email: catalin.stoean@inf.ucv.ro}
\date{}
\setcounter{Maxaffil}{0}
\renewcommand\Affilfont{\itshape\small}

\maketitle

\begin{abstract}

The Chaos Game Representation, a method for creating images from nucleotide sequences, is modified to make images from chunks of text documents. Machine learning methods are then applied to train classifiers based on authorship. Experiments are conducted on several benchmark data sets in English, including the widely used Federalist Papers, and one in Portuguese. Validation results for the trained classifiers are competitive with the best methods in prior literature. The methodology is also successfully applied for text categorization with encouraging results. One classifier method is moreover seen to hold promise for the task of digital fingerprinting.
\end{abstract}



\section{Introduction}

Every author has a specific style when writing a text, be it an email, a blog article, a paper, a book, and so on. Naturally, the larger the amount of text written, the more intense the style of the author is impregnated in the content. The author attribution (AA) task assumes the presence of several examples of documents that are written by various authors and it is desired to determine which one of them wrote a given anonymous text \citep{Holmes}. There have been numerous attempts to achieve such a goal, and according to \citep{Tweedie} the first one dating back to the 19th century, when the American physicist Thomas Mendenhall investigated the plays of Shakespeare \citep{Mendenhall}. His pioneering work relied on the word length, based on a suggestion by the logician Augustus de Morgan.

Since then, there have been many approaches to study the style (generally called stylistics \citep{Holmes94}). A part of stylistics is stylometry, which refers to the measurement (usually using computers) of style based on numerical analyses. The approaches that deal with stylometry include the use of certain words, their number of occurrences, the avoidance of some words, the richness of the used vocabulary, the study of the sentence structure and many others. An excellent overview of the existing AA methods can be found in \citep{Stamatatos}. The large use of the Internet has lead to a continuously growing volume of anonymous texts such as emails, blogs, forum messages, reviews for various products, source codes and so on, providing thus further test cases for AA. It is employed in a large variety of applications \citep{Stamatatos}, from the traditional literary research (attribution of anonymous literary works to known authors) \citep{Neme}, \citep{Hoover}, to intelligence (identification of the authors of messages from web forums to known terrorists) \citep{Alam}, \citep{Abbasi}, criminal and civil law (e.g. checking the authenticity of suicide notes or copyright disputes) \citep{Grant}, \citep{juola2013stylometry}, \citep{johnson2014identifying}, the detection of plagiarism \citep{Kuta}, program AA (determining the authors of malicious software source code) \citep{Layton}, \citep{Rosenblum}.

The current research proposes an original approach for AA that shifts from the traditional manner of considering the lexical, syntactic or semantic features. The proposed method transforms the alphabet into a binary representation. It then uses the Chaos Game Representation (CGR) to produce images that are subsequently classified via machine learning algorithms. The approach is validated on the Federalist Papers with various settings, but also on other data sets, including one in Portuguese, and the results are very competitive as compared to the traditional methodologies from the literature. Among the advantages of the proposed method, one is that, due to its inner mechanics, it can be directly applied on texts of any languages that use the Latin alphabet. Moreover there is no special treatment of any specific terms, and no need for a dictionary; the decisions are reached only on the images produced via CGR.

The rest of the article is organized as follows: the next section presents an overview of existing AA methods, as well as a brief description of the CGR. The latter will prove useful in section \ref{sec:method} where the proposed approach is presented in detail. Section \ref{sec:experiments} presents the results obtained when applying the methodology on several data sets under various settings and section \ref{sec:conclusions} encloses the article with some final remarks and ideas for future work.

\section{Prerequisites}

AA assumes the existence of a specific writing style for each author, with characteristics that the authors themselves are not aware of. Such a writing style acts as a $fingerprint$ \citep{Ebrahimpour}, as various features have been demonstrated to remain consistent for one author over the years \citep{Sayoud}. The proposed approach metaphorically achieves a representation that can be imagined as a fingerprint and subsequently, based on a training set, a machine learning classifier discovers the features of these depictions that distinguish between authors.

\subsection{The Federalist Papers}

Possibly the most influential work in AA was the book \citep{Mosteller} that brought to the general attention the Federalist Papers. These consist of 85 journalistic articles that were published between 1787 and 1788 in newspapers under the pseudonym $Publius$. Their initial aim was to persuade the New York citizens to ratify the American Constitution. There were three authors of the political essays, Alexander Hamilton, James
Madison and John Jay. Hamilton was the sole author of 51, Madison wrote alone 14, Jay was the sole author of 5 and 3 are produced from the joint contribution of Hamilton and Madison. There are still 12 items which are claimed by both Hamilton and Madison and these are generally referred to as the \textit{Disputed Papers}. Extensive research has been carried out to reach a consensus about the disputed papers and the conclusion is that Madison was likely the author of all 12 essays \citep{Tweedie}, \citep{Kjell}, \citep{HolmesForsyth}, \citep{LevitanArgamon}, \citep{Savoy}.

In the current work, the considered assignments are the ones from the Project Gutenberg 1.5 release of The Federalist Papers\footnote{\url{http://www.mirrorservice.org/sites/ftp.ibiblio.org/pub/docs/books/gutenberg/1/18/18.txt}}. This corpus is important and widely used because several criteria are met: the texts have a similar topic, genre and belong to the same time period. It is also regarded as nontrivial due to numerous observed similarities in the writing styles of the two authors under consideration.

\subsection{Overview of Previous Author Attribution Methods}

\begin{figure}
	\includegraphics[width=\textwidth]{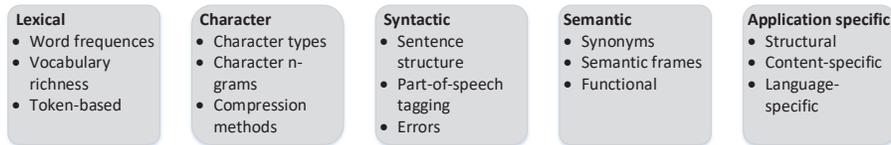}
	\caption{Classification of the stylometric features that are extracted from text for the AA task.}
	\label{fig:features}
\end{figure}

The AA is the task of identifying the author of a text from a group of several candidates based on text samples written by all these authors \citep{Stamatatos}, \citep{Zhang}. So there is a set of candidate authors, a training corpus comprised of a set of text samples for which the authors are known and belong to the initial set of candidates, and a test corpus that includes a set of text samples with unknown authorship. For each text sample in the test set the correct candidate author from the initial set has to be identified. In order to achieve this task, some relevant features should be extracted from the text samples. Figure \ref{fig:features} illustrates an enumeration of feature types that are usually considered for the AA problem \citep{Stamatatos}. After the feature extraction step, usually (but not always) feature selection is applied to keep only the most relevant attributes and subsequently a machine learning approach is employed to reach the document attribution.

Probably the most original step in achieving AA lies in the selection of the stylometric features. The first two categories in Figure \ref{fig:features} consider the text as a sequence of words and characters, respectively. Besides the counting of tokens that correspond to words, numbers or punctuation marks, lexical features may take into consideration other measures such as sentence length and word length counts, vocabulary richness (diversity of the vocabulary of a text) \citep{Neme} or the counting of the errors in a text (missing or mistakenly inserted letters, formatting errors etc) \citep{Koppel2003}. The most widely used approach is to obtain from each text a vector of word frequencies, that is the \textit{bag-of-words} representation that is common to text categorization, where each document has to be assigned to one or more categories. The \textit{function words} (e.g. prepositions, articles, pronouns etc.) are among the most valuable features when discerning between authors. This is in marked contrast to the problem of text categorization, where these words do not possess information about the categories and are usually referred to as \textit{stop words} \citep{Garcia}, \citep{Fung}. An approach that also proved to be successful is to identify the most frequent words that appear in the texts of each author \citep{Koppel}, \citep{Stamatatos2006}. Then, subsets of these words are retained to represent the characteristics of each author.

Measures like specific character count, upper or lower cases, number of digits or punctuation marks are often used within the character based features \citep{Zhang}. The next level feature processing regards the use of $n$-grams, i.e. each text is partitioned into overlapping pieces, each of character length $n$. These $n$-size chunks start from the first character, then from the second, third and so on, leading thus to a (nearly) $n$-fold increase over the size of the initial text. The $n$-grams usually take into consideration the space characters, punctuation marks etc. A problem that this approach faces is the choice of the proper value for $n$. While a larger $n$ helps to capture more contextual information and reduces redundancies, it also increases the memory needed to collect and retain such information (in the worst case, by a factor approaching $n$) \citep{Stamatatos}, \citep{Zhang}, \citep{Stamatatos2006}.

One class of methods that rely on character attributes involves compression-based approaches.
Texts are compressed via a dedicated tool, and the size-reduced representations thus obtained are then used for AA. A recent survey that focuses on such approaches for AA is described in \citep{Oliveira}.

The syntactic features refer to the manner of constructing sentences and phrases, since authors unconsciously follow specific patterns. The specific choices of nouns, verbs, adverbs etc., as well as their order in a sentence, their count or length has proven to be relevant for AA \citep{Stamatatos}, \citep{Zhang}, \citep{Halteren}. Erroneous use of verb tenses, sentences without any main clause, and run-on sentences represent only a few of the syntactic errors that were used in \citep{Koppel2003}; such information is also utilized by the human experts when faced with the AA task \citep{Stamatatos}.

The semantic features assume the association of certain words or sentences with semantic information. This also involves the use of synonyms or hypernyms \citep{Stamatatos}. Identification of the causal verbs is used in \citep{McCarthy}. Frame semantics were successfully used in \citep{Hedegaard} as markers for AA for translated texts.

The application-specific attributes refer to certain domains like emails, blogs, computer programs etc., where various specific keywords can be identified. Such particularities may refer to the manner of starting a message (greetings), signatures, certain HTML tags, abbreviations and so on \citep{Koppel2009}.

There have been many computational measures defined so far (several of which are mentioned above) and each has been shown to have a contribution in the identification of the authors, but they often work better when several are combined. The measures start from very simple ones like word length, sentence length to more sophisticated ones like the number of words that appear with a certain frequency, or the K-, S- and R-measures \citep{Koppel2009}. The Delta measure is derived from a more recent study \citep{Burrows}. It projects the documents to points in space and then assigns them to the author that is closest to them according to the proposed distance measure. This is a measure that was later largely used, especially in conjunction with other methods \citep{Hoover}, \citep{Garcia}, \citep{Jockers}. To the best of our knowledge, the CGR was only suggested as possibly being applicable to AA in \citep{Jeffrey} and in a conference talk \citep{Ramon}.

After the various attributes are extracted (or computed) from text, the usual steps are represented by feature selection followed by the application of a machine learning algorithm. This constructs classifiers from a training set, based on the correspondence between the texts and their known
authors. These in turn are employed in order to achieve the AA task on a given test set. A large variety of machine learning approaches have been successfully tried, like support vector machines \citep{Alam}, \citep{Ebrahimpour}, \citep{Fung}, decision trees \citep{Kim2011}, neural networks \citep{Tweedie}, \citep{Tsimboukakis2010}, ensemble classifiers \citep{Stamatatos2006} etc.

\subsection{Chaos Game Representation}

The CGR was proposed by Jeffrey \citep{Jeffrey}, \citep{Jeffrey92} as a means to visualize the structure of a DNA sequence. The representation starts from a square with the corners labeled as $A$, $C$, $G$ and $T$. The starting point is given by the middle point of the square. Next, the first nucleotide of the sequence is plotted as the middle point between the corner labeled by the current nucleotide and the starting point. This is the next current point and is used to represent the subsequent nucleotide and the process continues like this. If the obtained square image represented this way has a size of $2^k \times 2^k$ pixels, then it can be shown that every pixel represents a distinct $k$-mer \citep{Jeffrey}. The gray level of a given pixel is determined by the number of times its corresponding $k$-mer occurs in the DNA sequence, relative to the total number of $k$-mers.
The CGR images from the DNA sequences of various species illustrate distinct patterns such as triangles, rectangles or other complex fractal structures \citep{Wang}, \citep{Karamichalis2015}, \citep{Karamichalis2016}.

An alteration of the CGR procedure known as Frequency CGR (FCGR), proposed in \citep{Deschavanne}, is adopted in the current work. FCGR and CGR are equivalent once the pixelation level is fixed, but the original CGR form is not convenient to be processed by a computer, while FCGR is easier to implement (see \citep{Wang} for details). A $k$-th order FCGR of a sequence is a $2^k \times 2^k$ matrix that is achieved by splitting the CGR into a $2^k \times 2^k$ grid, defining the element $a_{ij}$ of the matrix as the number of points that are situated in the corresponding grid square. A first order and a second order FCGR are represented in equations (\ref{eq:FCGR1}) and (\ref{eq:FCGR2}), where $N_w$ is the number of occurrences of $w$ in sequence $s$.

\begin{equation}
FCGR_1(s) =
\begin{pmatrix}
N_C & N_G\\
N_A & N_T
\end{pmatrix}
\label{eq:FCGR1}
\end{equation}

\begin{equation}
FCGR_2(s) =
\begin{pmatrix}
N_{CC} & N_{GC} & N_{CG} & N_{GG}\\
N_{AC} & N_{TC} & N_{AG} & N_{TG}\\
N_{CA} & N_{GA} & N_{CT} & N_{GT}\\
N_{AA} & N_{TA} & N_{AT} & N_{TT}\\
\end{pmatrix}
\label{eq:FCGR2}
\end{equation}

Consequently, the $FCGR_{k+1}(s)$ will be defined by replacing each element $N_X$ in the previous $FCGR_{k}(s)$ with the four elements in (\ref{eq:FCGRk}) \citep{Wang}, \citep{Karamichalis2015}, \citep{Karamichalis2016}. The words with higher frequencies are displayed in the FCGR images as pixels with higher intensities: the darker the pixel, the higher is the frequency \citep{Deschavanne}.

\begin{equation}
FCGR_{k+1}(s) =
\begin{pmatrix}
N_{CX} & N_{GX}\\
N_{AX} & N_{TX}
\end{pmatrix}
\label{eq:FCGRk}
\end{equation}

\section{Proposed Methodology}
\label{sec:method}

As mentioned above, the FCGR was designed to work on an alphabet of only 4 letters. Some adjustments will be required in order to apply it for a text written in English. Firstly, all the text is transformed to lowercase. Next,
from the 26 letters plus digits and punctuation, a set of only 16 distinct characters is reached. Each such character
is then coded as a pair of two digits in base 4 and thus an alphabet of only 4 distinct characters is used (that is,
digits from 0 to 3). The base 4 characters are subsequently transformed into pairs of binary digits, for purposes of utilizing FCGR \citep{Deschavanne}, \citep{almeida_analysis_2001}, \citep{Wang}, \citep{Karamichalis2016}. Sequences of fixed length are then established and they are transformed into CGR images.
The images are separated into training, validation, and test sets. The training set is the input for
machine learning methods for classification, the validation set is used to tune, optimize, and assess the quality
of these classifiers, and finally they are applied to the test set.

\subsection{Reducing the Alphabet}

The canonical form of the FCGR requires an alphabet of four characters. In order to reach it by departing from the
Latin alphabet the most appropriate considered choice was that of the numerical representation in base 4. Still,
there are many ways in which a text could be transformed into a base 4 representation. The current choice is adopted
after several different attempts have been tried, and took into consideration several goals:
\begin{itemize}
	\item Obtain a base 4 representation that is not significantly larger than the original representation.
	\item Incur little or no information loss.
	\item If the choice is to use an alphabet of $4^2$=16 characters, find substitution rules that help to balance the frequencies of the 16 "equivalence classes" of characters.
\end{itemize}

\begin{figure}
	\includegraphics[width=\textwidth]{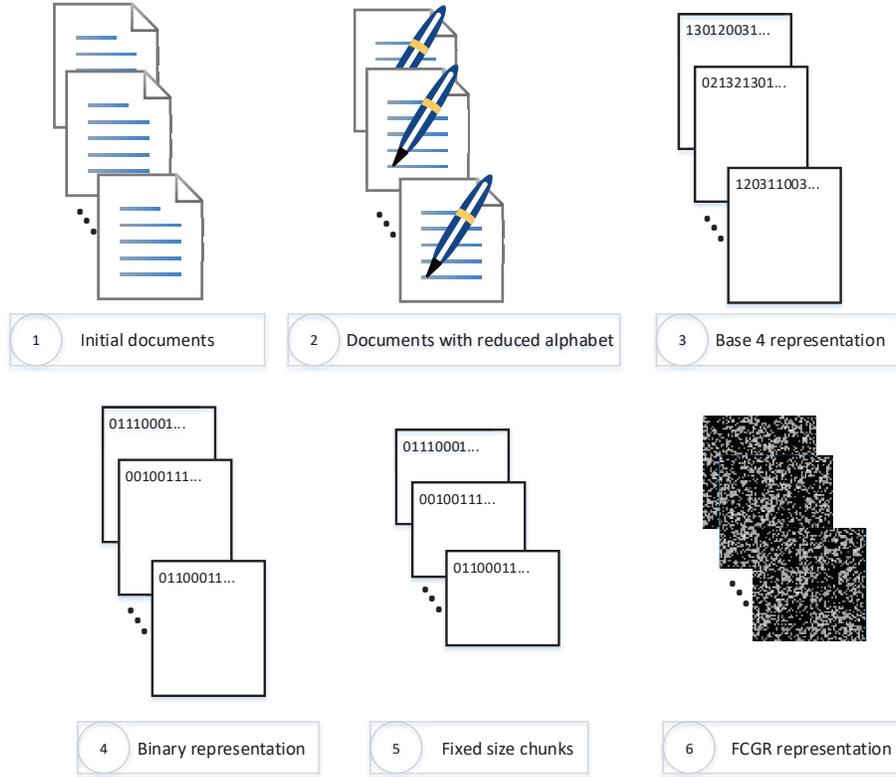}
	\caption{From text to FCGR. The alphabet is reduced to 16 characters and these are subsequently represented as pairs of numbers in base 4. Next, a binary representation is achieved and sequences of fixed size are considered for producing the FCGR images.}
	\label{fig:framework}
\end{figure}

An intuitive representation of the text to FCGR image transformation is presented in Figure \ref{fig:framework}.

The most successful representation among the ones tried is obtained by encoding each distinct
character as a pair of numbers in base 4. There are only 16 different pairs of numbers in base 4, so some characters are replaced by others in the initial text in order to reach only 16 equivalence classes. Table \ref{tab:encoding} illustrates the chosen association of characters and the selected encoding as pair of numbers in base 4. The classes of characters are chosen and ordered so that several criteria are met:
\begin{itemize}
	\item The letters that are very common are left alone in order to reach a rough balance with respect to the
	occurrence of each equivalence class of characters.
	\item The letters that are less common are grouped in the same equivalence class if they have similar sounds or similar phonetic usage.
	\item White space (tabs, spaces, and line breaks) are grouped into one equivalence class.
	\item The punctuation and the digits are gathered in the same class.
	\item Other characters (that do not belong to the Latin alphabet) are omitted.
\end{itemize}

\begin{figure}
	\begin{tabular}{c}
		\includegraphics[width=0.79\textwidth]{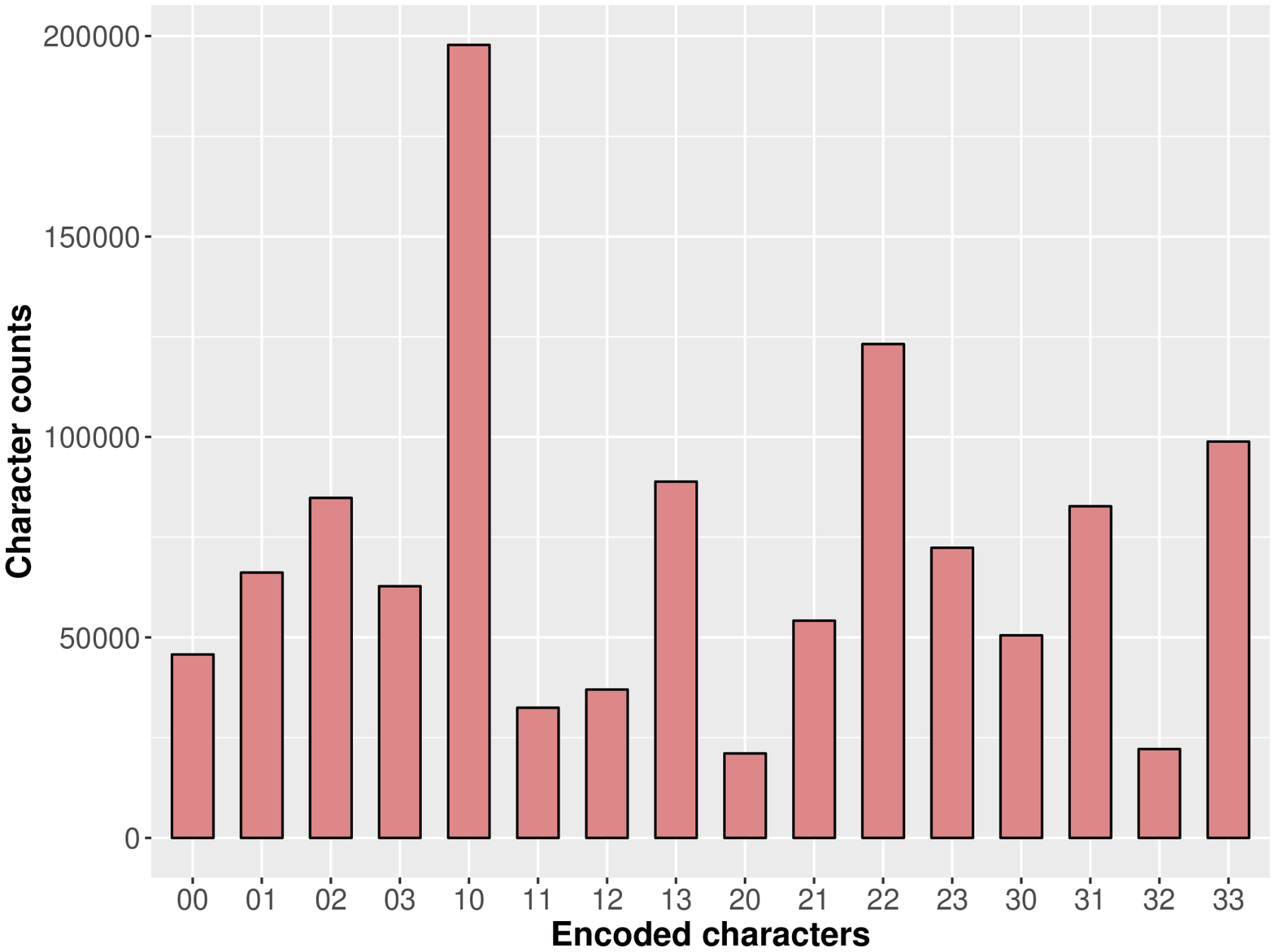} \\
		\includegraphics[width=0.79\textwidth]{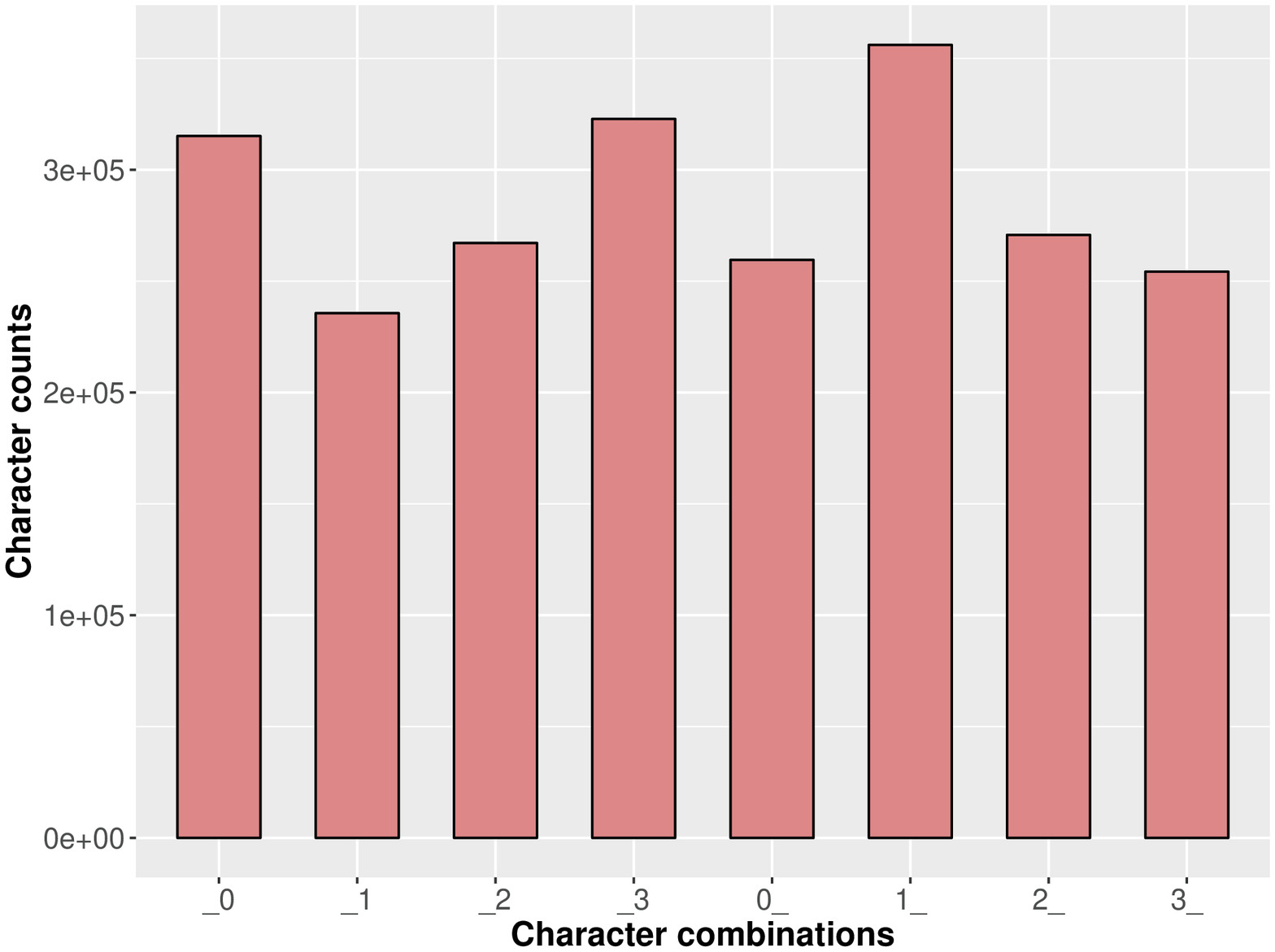} \\
	\end{tabular}
	\caption{The top plot illustrates the number of counts for the base 4 encoding of the characters (see Table~\ref{tab:encoding}). The bottom one shows the distribution of characters that start and end with each of the four digits. The counts refer to the occurrences of the characters in the Federalist Papers.}
	\label{fig:counts}
\end{figure}

In order to test the combinations, tests were done on the Federalist Papers. The first plot in Figure~\ref{fig:counts} shows how the occurrences of the encoded digits are distributed. As expected, the space character (encoded as 10) is the one that is the most common, while the digits equivalence class (which also includes the punctuation marks), and the the $p$ class, are the two least frequent.

\begin{table}
	\caption{The equivalence classes and their corresponding base 4 representation. \label{tab:encoding}}
	\begin{tabular}{|c|c|c|c|}
		\hline
		Class & Base 4 &  Class & Base 4 \\
		\hline
		\{b, d\} & 00 & \{p\} & 20 \\
		\{a\} & 01 & \{r\} & 21 \\
		\{i, y\} & 02 & \{e\} & 22 \\
		\{h, j, g\} & 03 & \{o\} & 23 \\
		\{\textvisiblespace, $\backslash$t, $\backslash$n \} & 10 & \{u, v, w\} & 30 \\
		\{l\} & 11 & \{s\} & 31 \\
		\{c, k, q, x, z\} & 12 & \{(, ), -, +, [, ], 0, 1, ..., 9, ?, !, :, ;, ,, .\} & 32 \\
		\{m, n\} & 13 & \{t\} & 33 \\
		\hline
	\end{tabular}
\end{table}

Beside the grouping of the characters, special attention is paid to the choice of the base 4 representation for each equivalence class. It is intended to have a good balance in the spread of characters that are encoded in base 4 and start, and end respectively, with a certain digit. The encoding in Table \ref{tab:encoding} leads to a representation that has a good balance, as illustrated in the second plot of Figure \ref{fig:counts}.

\subsection{From Characters to FCGR}

After the text is transformed in a base 4 representation, as described in the previous subsection, the FCGR will be computed. In order to get to the representation, pairs of binary digits are subsequently needed \citep{Deschavanne}, \citep{almeida_analysis_2001}, \citep{Wang}, \citep{Karamichalis2016}. Every binary sequence that corresponds to a document can be used to create a FCGR image.

In most experiments long binary sequences are cut into fixed length chunks, with a different FCGR image created from each such chunk. Exceptions are made in cases where the strings may not be sufficiently long to generate multiple images. Some experiments are devoted to finding "good" chunk lengths for the purpose at hand.

Depending on the established binary sequence length, there are several binary chunks that can be obtained from a document and for each one of them a distinct FCGR is obtained. In order to have text chunks of the same size, the "tail" sequence, which is smaller than the predefined length, is discarded. For both training and validation sets, the association of each chunk
(and obtained FCGR) with the corresponding author of the document from which the chunk was taken is recorded for later use.

\subsection{Image Classification}
\label{subsec:image_class}

The training set, initially holding $n$ text documents is further on represented by a set
of (when chunks are used, possibly more than $n$) FCGR images. Each such image represents the $signature$ of the author
who wrote the document from which the FCGR image was created.
The images from the test set are treated similarly: for each image an author is identified, after which these computed labels are used to establish the author of the entire document by weighted vote.

Three machine learning techniques are applied in turn for classifying the FCGR image by author correspondence. Two of the methods are state-of-the-art classifiers, namely support vector machines (SVM) \citep{vapnik} and logistic regression (LR) \citep{freedman}. The parameter settings of the two approaches are kept to default values. Neural networks, random forests and nearest neighbors have also been tried within the pre-experimental stage but with poor results. The Fourier Trig Transform with Principal Component Analysis (FTT+PCA)
is the third employed methodology. This is a variant of the recently proposed method \citep{lichtblauSYNASC16} and thus some details about it are further provided. Either a Fourier sine or cosine transform of each FCGR image is computed. For some value $k$ the lowest $k \times k$ dimensional array of frequencies is flattened into a single vector. From the matrix of such vectors (one for each training image) and another value $n$, the $n$ largest principal components extracted from it are retained along with the matrices that provide the corresponding left and right singular vectors. The scaled left singular vectors are stored in a kD tree, while the right singular vectors are required for preprocessing the lookup images. For a given preprocessed
image, the nearest neighbors found in the kD tree are weighted, closest to furthest, by the inverse of the Euclidean distance from query vector to neighbor. These weights are re-scaled so that their sum equals unity.
This is done in order to have values that would resemble probabilities reflecting the
degree of belonging to the classes (in this case, the authors). Besides \citep{lichtblauSYNASC16}, previous
tandems of Fourier and PCA techniques were realized by \citep{AshokRajan2010} and \citep{Zhangetal2013}. The
originality of the current FTT-PCA version comes from the computational efficiency, a paring-down of the
implementation, and also improved quality of results vs not using it.

\section{Experimental Results}
\label{sec:experiments}

The proposed methodology is tested on several English data sets and one Portuguese. The proof of concept is realized on a collection of various English text documents and they are briefly introduced in the first experiment. Two settings are investigated in the second experiment on the Federalist Papers: chunk length and pixelation level. Good settings for both are found. They are used in the third experiment, which covers three benchmark data sets from the AA literature. The last experiment uses a Portuguese language data set from Brazilian newspaper articles (100 authors, 30 articles from each, spanning 10 genres). In addition to authorship attribution, an exercise in text categorization on that data set is presented.

\subsection{Experiment 1: Proof of Concept}

\begin{figure}
	\centering
	\begin{tabular}{cc}
		\includegraphics[width=0.49\textwidth]{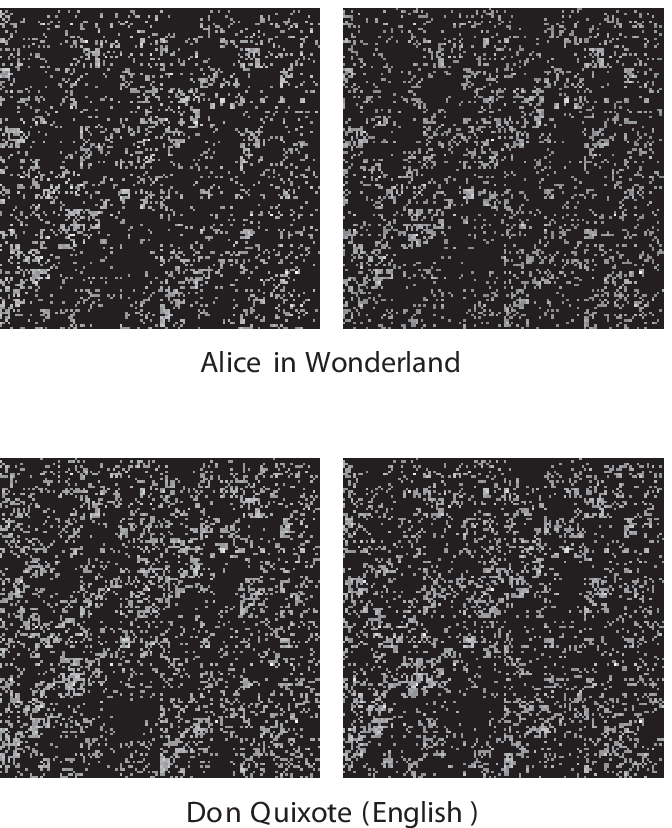} &
		\includegraphics[width=0.49\textwidth]{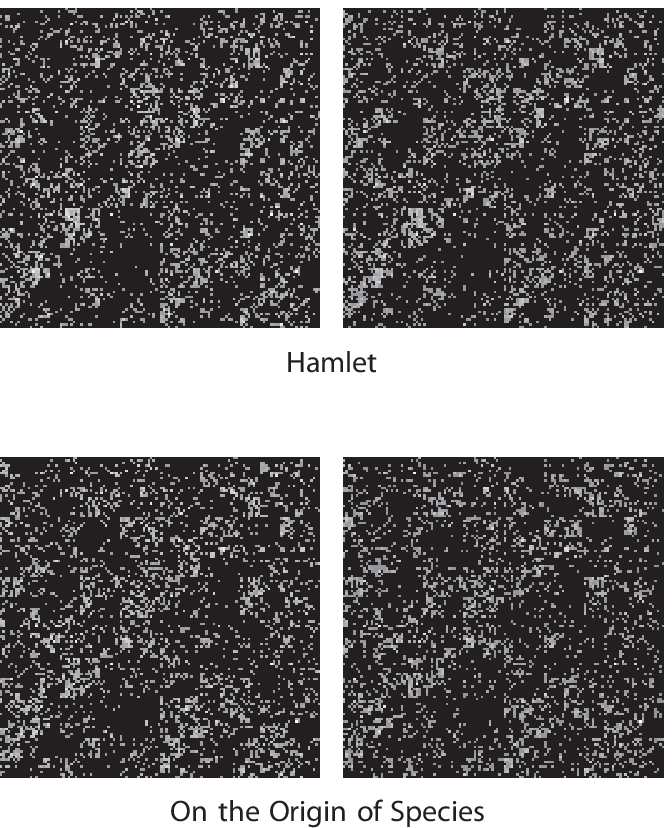} \\
	\end{tabular}
	\caption{FCGR representation of the first and last chunks from various texts.}
	\label{fig:samples}
\end{figure}

The idea of the methodology was initially tested using 16 English texts that are taken directly from the Mathematica software. These are enumerated in Table \ref{tab:booksMathematica} together with the number of text chunks of a fixed size of 8500 base 4 characters. The number of text chunks is indicated with the only purpose of illustrating the size of each manuscript. All the considered documents have at least 17000 characters so as to contain minimally two chunks per text. Figure \ref{fig:samples} illustrates some samples obtained through the proposed FCGR methodology. While for the human eye the differences are almost nonexistent, the machine distinguishes between them remarkably well.

\begin{table}
	\caption{Text documents considered for the first experiment and the corresponding number of chunks with a size of 8500 base 4 characters. \label{tab:booksMathematica}}
	\centering
	\begin{tabular}{|c|c||c|c|}
		\hline
		Text name & Chunks & Text name & Chunks\\
		\hline
		Aeneid & 140 & Alice in Wonderland & 11\\
		Beowulf Modern & 30 & Code of Hammurabi & 12\\
		Don Quixote & 241 & Federalist 10 & 4\\
		Genesis KJV & 48 & Hamlet & 40\\
		Magna Carta & 6 & On the Nature of Things & 97\\
		Origin of Species & 210 & Plato's Meno & 16\\
		Pride and Prejudice & 159 & Shakespeares Sonnets & 22\\
		UN Human Rights & 2 & US Constitution & 11\\
		\hline
	\end{tabular}
\end{table}

This is intended as a brief "proof of concept" and FTT+PCA is the only method applied for this particular test. Each document has a distinct author and the goal is to show that similarity is considerably greater between parts of the same document than between parts of different documents. In particular, for each chunk, a search is conducted for the most similar chunks from amongst all documents in the set. From the 16 initial texts, 1049 text chunks in total are produced. Each such part is then used to produce a FCGR image. FTT+PCA  is used to find, for each image, the closest 3 neighbors from the entire set (but excluding the search image). Out of 1049 images, 1046 have all three nearest neighbors coming from chunks within the same document as the respective test chunk. Another has two closest neighbors out of three coming from chunks in the same document. The remaining two each have the nearest chunk coming from the same document. These are in fact the two chunks from the \textit{UN Human Rights}, so in this case the outcome is optimal since all other neighbors of necessity come from different texts. These primary results encouraged subsequent, more thorough experiments, conducted on benchmark data sets. They provide further evidence that the method under study is powerful and also help to better understand ranges of useful settings for this methodology.

\subsection{Experiment 2: Federalist Papers}

The first important test herein is on authorship of the Federalist Papers. This will include a search for good parameter settings to use, as assessed by validation. There are two validation sets considered: One is obtained by taking chunks from the documents in the training set, e.g. the final FCGR image obtained for each document from the training set, while the other is comprised of FCGRs obtained from documents of known authorship, that were withheld from the training set. More specifically, the latter validation set is comprised of all chunks from 3 randomly selected essays by Madison and 8 randomly selected essays by Hamilton. Naturally, there is no FCGR image that appears in any two sets, so all the four sets (training, test and the two validation ones) are disjoint. The test set comprises the disputed essays, as this represents a common practice \citep{Stamatatos}, \citep{Ebrahimpour}. For purposes of testing, we use the modern consensus that the correct author for all disputed essays is Madison. The training and validation sets are the same for each classifier in order to observe the differences in results. As the individual documents are generally large, this allows for the separation into text chunks of various sizes. The size (pixelation level) of the squared FCGR images provides another parameter to be set. The focus of the current experiment is thus represented by the two parameters, text chunk size and pixelation level.

It is natural for any classification technique in general that the larger the input documents are, the more information is extracted and therefore the more accurate the results are. However, it is often the case that the documents to be evaluated are not large but, on the contrary, they are very small, so it is important to identify an author of relatively small pieces of texts. It is thus useful to learn how well this approach might work on relatively small texts, as well as assess what might be a good chunk size to use when texts are reasonably large. The settings in this experiment for the chunk sizes range from 500 to 10000 base 4 characters, in step sizes of 500. In the original text, that corresponds to ranging from 250 Latin alphabet characters to 5000 in step sizes of 250. In order to decide the label for an essay, the probabilities for each possible author are summed over all chunks in that essay. The maximal value determines the attributed author.

As concerns the second parameter that is investigated, the size of the squared $2^k\times2^k$ FCGR images, the values for $k$ are varied between 4 and 7. Smaller values make the image small, faster to process, but lighter, as the gray level of the pixels is proportional to the total number of $k$-mers. On the other hand, images produced when $k$ is larger contain more refined information, but the runtime and the necessary memory resources are increased. For FTT+PCA only pixelation levels of 6 and 7 are considered since smaller values do not support the number of Fourier components needed to get viable results.

The focus in the first part of this experiment is on finding reliable parameter settings. So the
classification accuracy is not calculated for the test set with the disputed essays
(where there is only a consensus opinion), but rather on the validation sets,
for which there is established ground truth.

\begin{figure}
	\begin{tabular}{cc}
		\includegraphics[width=0.49\textwidth]{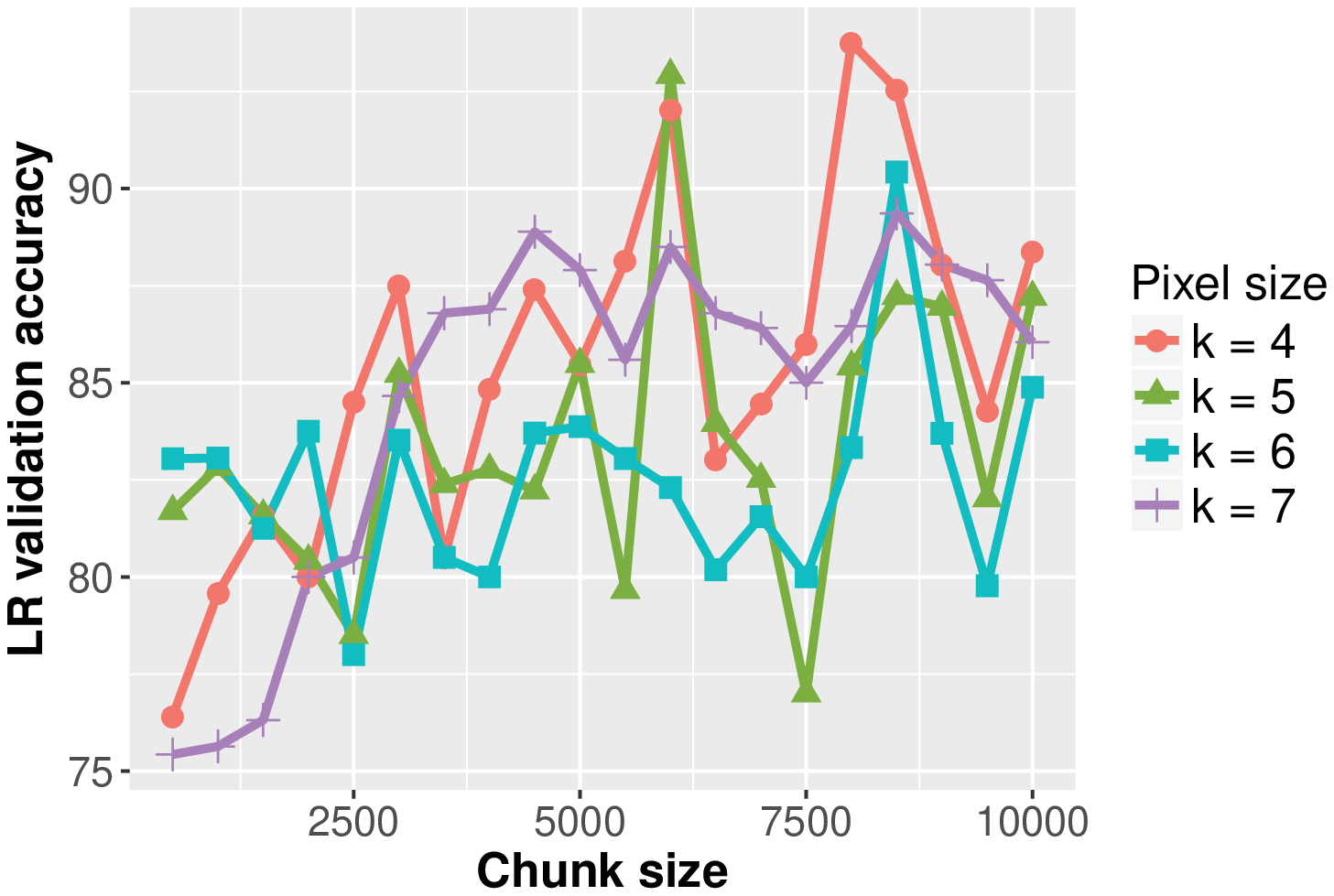} &
		\includegraphics[width=0.49\textwidth]{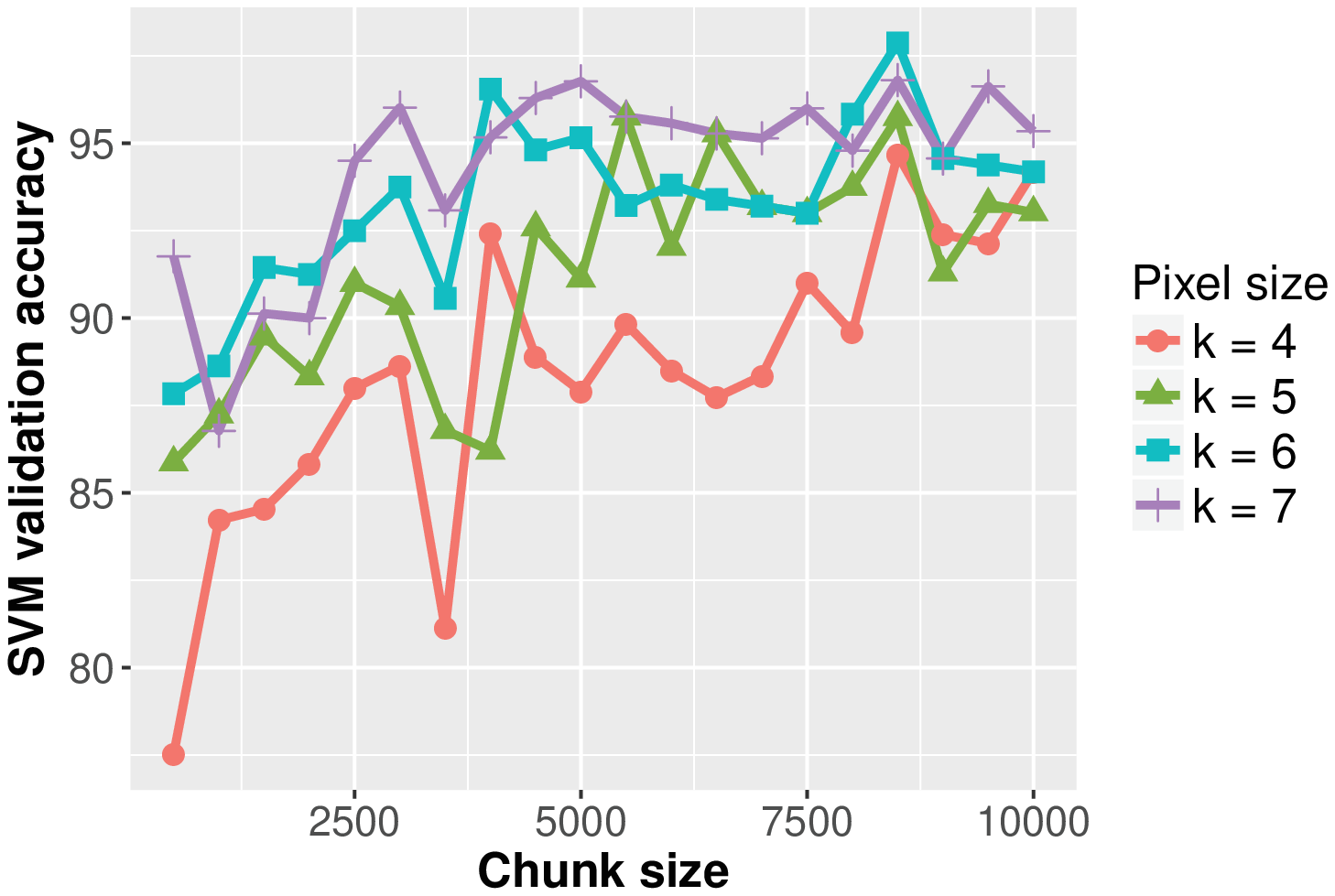} \\
		\includegraphics[width=0.49\textwidth]{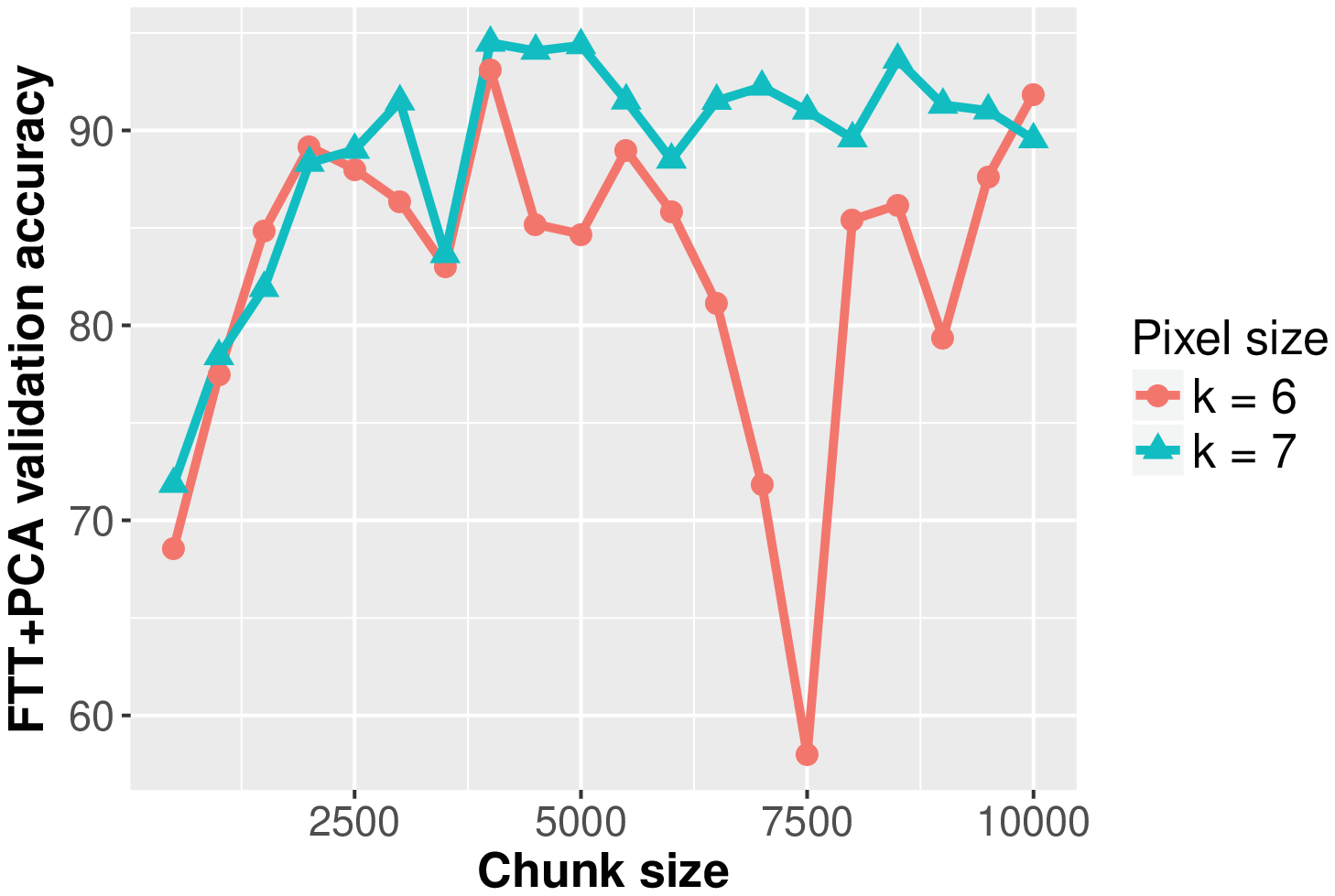} &
	\end{tabular}
	\caption{Classification accuracies obtained on the validation sets as obtained by the LR, SVM and FTT for chunk sizes varied from 500 to $10^4$ base 4 characters. For LR and SVM the size of the squared FCGR image is considered from k=4 up to k=7, while for FTT only the most promising two are illustrated, i.e. $k$=6 and $k$=7. Larger chunk sizes and higher pixelation level (k=7) assures more robust results.}
	\label{fig:pixChunks}
\end{figure}

Figure \ref{fig:pixChunks} illustrates the results obtained for the discussed settings for LR and SVM in the plots on the first row. It can be observed that for both methods the results are consistent, as the deviations remain low for $k$=7, and for SVM also for $k$=6. The FTT+PCA plot (the second row from the same Figure \ref{fig:pixChunks}) indicates that $k$=6 appears less stable than $k$=7, and this is also seen, though to a smaller extent, for the LR method. There had been pre-experimentation with $k$=8, which gives images that are larger and thus can show more detail. This setting was abandoned due to significantly increased run time for training, an increase on memory requirements, and essentially no real improvement observed in outcomes as compared to $k$=7.
As concerns the chunk size, a value at which results tend to stabilize is 3000 and on average over the three methods for $k$=7 the highest accuracy is achieved for 8500.

Accordingly, the parameter settings of chunk size 8500 and $k$=7 are used for the subsequent test on the Federalist Papers, while randomly varying the training and validation samples in 30 different repeated runs. The aim is to ascertain that the methods are not sensitive to the choice of used training and validation sets. Table \ref{tab:randomFeds} illustrates the correctness percentages for the two validation sets and for the test set, respectively, for all three classifiers. It can be observed that all of them have higher classification accuracy on the first validation set as compared to the second one: the former is obtained by gathering the FCGR images produced from the final chunk of each essay from the training set, therefore it is natural that those text parts resemble more the rest of the essays that remained in the training sets. As concerns the test set, in all trials, 10 of the 12 essays are attributed to Madison by both classifiers. The Federalist Papers 62 and 63 are the ones that, in some trials, get attributed to Hamilton. For LR this happens in 2 of the 30 trials for essay 62, and in 1 trial for essay 63. For SVM the situation is quite different: in 24 times of the 30 trials the essay 62 is attributed to Hamilton, and in 25 trials the same situation occurs for essay 63. The results of FTT+PCA are more similar to those of the SVM: in 27 of 30 trials it assigns essay 62 to Hamilton and in 29 cases essay 63 is labeled also with Hamilton. As concerns the other disputed essays, there is only one minor exception for essay 50, which is assigned by FTT+PCA in 1 trial out of 30 to Hamilton. When taking into account the current consensus, that all disputed essays are written by Madison, and summing up all trials for each classifier, the test accuracies are the ones from the final column in Table \ref{tab:randomFeds}. Nevertheless, \citep{Mosteller} claims that papers 62 and 63 are the ones "historians feel weakest about", so it can be observed that the results of the proposed methodology get complicated for the exact essays that are considered tricky.

\begin{table}
	\caption{Average accuracy results for the classification on validation and test sets over 30 trials, each with random samples in the training and validation sets. The test set comprises the disputed essays and the ground truth is considered with Madison as author. \label{tab:randomFeds}}
	\centering
	\begin{tabular}{|c|c|c|c|}
		\hline
		Classifier & Validation set 1 (\%) & Validation set 2 (\%) & Test set (\%)\\ \hline
		LR & 94.8 & 92.3 & 99.17\\
		SVM & 95.9 & 93.7 & 83.39\\
		FTT+PCA & 94.9 & 90.2 & 84.17\\
		\hline
	\end{tabular}
\end{table}


Based on the current experiment, it can be concluded that a minimum length for a document (or text chunk) should be of 2000 base 4 digits, but results improve considerably from 4500. To crude approximation, "a thousand words is worth a picture".

\subsection{Experiment 3: Tests on Large English Data Sets}

Two corpora are considered next, CCAT and PAN-12. CCAT is a subset of the Reuters Corpus Volume 1 \citep{Lewis} and despite the fact that it was initially meant for text categorization, there are several studies that used it for authorship attribution \citep{STAMATATOS2008}, \citep{Sapkota2015}. Two subsets are used, CCAT-10 and CCAT-50. As their names might suggest, the first one has 10 authors and the second one has 50. Each author has 100 articles, split into a set of 50 for training and 50 for testing (so there is no randomizing of the data in these experiments). Since the texts are already relatively short, no chunks are used for this data set. The general topic of the documents is corporate/industrial news.

PAN-12 represents a data set that was put forward in a workshop competition \citep{Juola12}. The current study uses problem J, where the genre is science fiction, and this is regarded in \citep{Juola12} as the most difficult of the AA challenges in this particular competition. The training set comprises documents of the sizes of novels, from $4\cdot10^4$ to $17\cdot10^4$ words, with 14 authors represented. The test set comprises 16 novels, one per candidate author and 2 that are not assigned to any of the 14 authors, so they should be labeled "none of the above" (NoA). Some thresholding is required in order to manage the NoA label within the proposed approach. Specifically, a test sample is assigned
to a certain author only if the probability for that author is of at least 0.3 (and of course is highest),
and it exceeds the second highest probability by at least a factor of 1.5. Chunks of text of 8500 base 4 characters are used for PAN-12 data set.

The same $k$=7 is used for the data sets in the current experiment and LR is kept as the subsequent classifier for the FCGR images, especially because the data sets have high number of classes (i.e. authors) and SVM tends to be considerably slower for these cases.

\begin{table}
	\small
	\caption{Results of the proposed technique with the parameter settings established within the previous experiment and of other methods on the considered benchmark data sets. \label{tab:ComparedResults}}
	\centering
	\begin{tabular}{|l|c|c|c|}
		\hline
		Method & CCAT-10(\%) & CCAT-50(\%) & PAN-12 \\ \hline
		FCGR+LR & 82.2 & 70 & 14/16 \\ \hline
		SVM & \multirow{2}{*}{86.4} & \multirow{2}{*}{-} & \multirow{2}{*}{-} \\
		\citep{Escalante} & & &\\ \hline
		SVM \citep{Sapkota2015} & 78.8 & 69.3 & - \\ \hline
		n-gram char (1,2) & \multirow{2}{*}{77.8} & \multirow{2}{*}{70.16} & \multirow{2}{*}{-} \\
		\citep{Sari} & & &\\ \hline
		n-gram char (2,3,4) & \multirow{2}{*}{74.8} & \multirow{2}{*}{72.6} & \multirow{2}{*}{-} \\
		\citep{Sari} & & &\\ \hline		
		SVM \citep{Plakias2008} & 80.8 & - & - \\ \hline
		Character n-grams \citep{Stamatatos2017} & 80.6 & - & - \\ \hline
		MSMF+FLF \citep{Sapkota2013} & 78.8 & 69.5 & - \\ \hline
		Best PAN-12 \citep{Juola12} & - & - & 14/16\\ \hline
		Average PAN-12 \citep{Juola12} & - & - & 10.8/16\\
		\hline
	\end{tabular}
\end{table}

Table~\ref{tab:ComparedResults} illustrates on the first row the results of the proposed methodology for the 3 data sets. The other rows contain results by various techniques that were applied for the same instances. The best reported result for CCAT-10 is obtained by SVM with bag of local histogram \citep{Escalante}. Subsequent work suggests that this particular result may be fragile insofar as it relies on parameter settings that are difficult to discern. There are two separate attempts to replicate their method, \citep{Sari} and \citep{Potthast2016}, and while they did reasonably well, the correctness assessments are notably lower, at 77\% and 75.4\% respectively. The next best result for this data set is achieved by the proposed methodology. In order to make our results reproducible, the source code is made available at the webpage \url{https://github.com/catalinstoean/FCGR-LR}. For CCAT-50, the proposed method achieves competitive results as compared to the ones found in the literature.

For the PAN-12 data set, there were 20 approaches that entered the competition for the considered benchmark. Out of these, there was only one to hit 14 out of 16 correct answers. The weakest reached 6 correct answers, while the average over all techniques was of around 11. Using LR, and aggregating chunk scores as in earlier experiments, 14 are correctly identified. The remaining 2 were by authors 6 and 11 respectively and were both placed in the NoA category (along with the 2 that actually were NoA). Obviously, the prior information that each author appears exactly once in the correct assessment is not used. It should be noted that author 6 did show up as second highest score in that particular wrong result. The incorrect result for author 11 is perhaps more interesting int hat the test turns out to be quite difficult. The novel in question is "Ripping Time", coauthored by Linda Evans and Robert Asprin. The training novels were "Mything Persons" and "Myth Inc. in Action", both by Asprin alone.

As the documents from the PAN-12 data set are significantly larger than the other benchmark ones used, it was decided to also vary the sizes for the chunks of text to larger values, from $10^4$ to 5*$10^4$ in increments of 5000. The results varied from 13 to 15 correctly attributed novels, with the best result found for 25000 characters.


\subsection{Experiment 4: Portuguese Data Set}

The next test for this methodology comes from a set of 100 authors of newspaper articles in Brazil \citep{VarelaJustinoOliveira}. There are 10 separate genres, each represented by 10 authors, each with 30 articles (although a few contained duplicates, after removal of which some authors had only 28 or 29 distinct articles). This is a fairly large test set and serves to show that the methods scale reasonably well to a larger set of authors. As the language is Portuguese rather than English, a modification was made in handling the alphabet: all diacritical marks were removed. Character substitutions then proceeded as in the other experiments. Also some articles were quite short and so no chunk size was used. Instead each image is created from an entire article, regardless of its length. As images are normalized so that the largest pixel value is 1, this is not a major departure although it does perhaps confer a modest bias toward recognition of authorship in cases where a particular author happens to write several articles of similar length. Many articles begin with headers that may contain common information e.g. publication and/or author name. The first 50 characters from each article are thus removed so as to make certain this introduces no bias into the tests.

The authors of \citep{VarelaJustinoOliveira} use 60\% of the articles for training and validation, and 40\% for testing. In order to achieve a similar split, and taking into account duplicates, the test set below is constructed as follows. From each block of five articles, the second and fourth are withheld for testing, except for the sixth block where instead the second and third are kept out. All articles not held out are used for training. In total there are 2990 articles, of which 1200 are put into the test set and the rest into the training set. Images were created at the pixelation level of 7, that is, $2^7$ x $2^7$ pixels. FTT+PCA and LR are used as training methods for this data set.

\begin{table}
	\small
	\caption{Results in percents for AA and genre classification for the Brazilian newspaper data set. The results of the proposed methods using LR and FTT+PCA options are presented beside results presented by other methods in the literature. For the proposed methods, percents for cases when the correct author is second and 3rd or 4th respectively are also shown.
		\label{tab:brazilian}}
	\centering
	\begin{tabular}{|l|c|c|c||c|}
		\hline
		Chosen & Correct & Next  & 3rd or 4th & Genre\\
		classifier & author & author & author & classification\\ \hline
		FCGR+LR & 82 & 6.9 & 4.1 & 86.8 \\ \hline
		FCGR+FTT+PCA & 63.3 & 12.2 & 8.8 & 82.5 \\ \hline
		SVM \citep{VarelaJustinoOliveira} & 72 & - & - & 86 \\ \hline
		SVM \citep{Oliveira} & 77 & - & - & 80 \\
		\hline
	\end{tabular}
\end{table}

The testing proceeds as follows. The FTT+PCA method had settings to retain the lowest 100x100 submatrix of Fourier frequencies, and the largest 28 singular values and corresponding right singular vectors. For each processed text image, the 12 closest neighbors are found. A score for the classifier is constructed, as before, based on reciprocal distances between neighbors and test vector. The prospective author with top score is the main guess. In 63.3\% of the tests this guess is in fact the correct author. Another 12.2\% have the correct author as second highest scoring prospective author. Nearly 9\% of the correct authors are found amongst the third and fourth highest scoring prospective authors. Loosely, nearly 85\% of the test cases have authors in the top four guesses. The results for FTT+PCA, as well as those for LR and two other methods taken for comparison are illustrated in Table \ref{tab:brazilian}. As it can be observed, for the LR method the results are even better. Again each test can be ascribed to any of 100 different authors, and this classifier ascribes an internal probability to each author for a given test image. In 82\% of the test cases the correct author is in fact the one that the classifier has identified as having the highest probability. Moreover, in 93\% of the tests, the correct author lies amongst the top four candidates.

Similarly to \citep{Oliveira} and \citep{VarelaJustinoOliveira}, these two tests were repeated with the goal of classifying genre rather than specific author (see right column in Table \ref{tab:brazilian}). Not surprisingly the correctness percentages go up. With the FTT+PCA method the correct genre is identified in 82.5\% of the tests, with the second guess being the correct genre in another 9.25\% of the test cases. LR correctly identifies genre in 86.8\% of the tests, with another 9.2\% having the correct genre as second best guess.

The results indicated above compare favorably with prior work involving this data set. The compared results are chosen from the methods that performed best in \citep{VarelaJustinoOliveira}, \citep{Oliveira}.

It should be remarked that the FTT+PCA method, while not producing the best results, shows an interesting possibility. It is fairly fast, as train and test vectors, after preprocessing, are vectors of 28 double-precision values. These could be used as a "signature" (that is, a digital fingerprint) for purposes of testing for possible plagiarism. If one tests against a known body (the "training" set) and finds one or more close neighbors one might then use more expensive tools to assess whether plagiarism is likely. So this method might have potential either as a preprocessor, or as a cheap secondary system, say to help corroborate the more commonly used methods for this task.

\section{Conclusions}
\label{sec:conclusions}

The proposed methodology uses the frequency chaos game representation to produce grayscale images from text. The images are subsequently used to train machine learning classifiers and the learned models can identify with a high accuracy new texts that are also represented by such images. There are several means of transforming the text into images and various settings are tested. For the generation of images, the number of distinct characters was reduced to 16, since a small power of 4 is desirable for this methodology. Moreover, choices are made for character substitutions that tend to balance sizes of the equivalence classes. Similarity of sound or phonetic usage is  also considered in creating these equivalence classes. Several sets of equivalence classes were considered, with the one that seemed to work best being shown herein. A possible future direction might be to employ a metaheuristic to search for an optimal distribution of the characters into equivalence classes, also taking into account particularities from different languages.

The methodology gives compelling results for the corpora considered both in English and Portuguese. The validation results on both the Federalist Papers and the Portuguese data set are quite competitive with the best in the literature. The latter furthermore indicates that the method might hold promise for identification of genre as well as authorship.

One method described, FTT+PCA, could be also used to provide "signatures" for texts. This can be useful for detecting possible plagiarism, for example. Moreover, as this method delivers reasonable results even for texts as small as a thousand or so characters, it might be applied to author identification when analysing anonymous emails or blog articles. These are interesting possibilities for further research.

\section*{Acknowledgements}

 The authors thank Ruxandra Stoean for proof reading and making several comments that helped to improve the exposition. Daniel Lichtblau thanks Jakub Kabala for introducing him to the topic of authorship attribution. He thanks Hans Hock for a very helpful discussion on linguistic aspects of character substitution as they relate to this topic.

\section*{References}
\bibliographystyle{apalike}
\bibliography{DLCS}

\theendnotes

\end{document}